\documentclass{article} 
\usepackage{iclr2024_conference,times}
\usepackage{graphicx} 

\usepackage{amsmath,amsfonts,bm}

\def\eqref#1{equation~\ref{#1}}

\def\1{\bm{1}}

\DeclareMathAlphabet{\mathsfit}{\encodingdefault}{\sfdefault}{m}{sl}
\SetMathAlphabet{\mathsfit}{bold}{\encodingdefault}{\sfdefault}{bx}{n}

\usepackage{hyperref}
\usepackage{url}
\usepackage{algorithm}
\usepackage{algorithmic}
\usepackage{mathtools}
\usepackage{etoolbox}
\usepackage{booktabs}[demo]
\usepackage{wrapfig,lipsum}
\usepackage[font={small}]{caption}
\definecolor{ForestGreen}{RGB}{34,139,34}
\usepackage{wrapfig}
\usepackage{wrapfig,lipsum,booktabs}

\title{Oh! We Freeze: Improving Quantized \\ Knowledge Distillation via Signal Propagation Analysis for Large Language Models}

\author{Kartikeya Bhardwaj\thanks{Equal Contribution. $^\dag$Qualcomm AI Research is an initiative of Qualcomm Technologies Inc.}\ , \  Nilesh Prasad Pandey$^*$, Sweta Priyadarshi, \textbf{Kyunggeun Lee},\\ \textbf{Jun Ma}, \textbf{Harris Teague}\\
Qualcomm AI Research$^\dag$  \\
\texttt{\{kbhardwa,nileshpr,swetpriy,kyunggeu,jma,hteague\}@qti.qualcomm.com}}

\iclrfinalcopy 
\begin{document}

\maketitle
\vspace{-0.2in}
\begin{abstract}
Large generative models such as large language models (LLMs) and diffusion models have revolutionized the fields of NLP and computer vision respectively. However, their slow inference, high computation and memory requirement makes it challenging to deploy them on edge devices. In this study, we propose a light-weight quantization aware fine tuning technique using knowledge distillation (KD-QAT) to improve the performance of 4-bit weight quantized LLMs using commonly available datasets to realize a popular language use case,  on device chat applications. To improve this paradigm of finetuning, as main contributions, we provide insights into stability of KD-QAT by empirically studying the gradient propagation during training to better understand the vulnerabilities of KD-QAT based approaches to low-bit quantization errors. Based on our insights, we propose \textit{ov-freeze}, a simple technique to stabilize the KD-QAT process. Finally, we experiment with the popular 7B LLaMAv2-Chat model at 4-bit quantization level and demonstrate that ov-freeze results in near floating point precision performance, i.e., less than 0.7\% loss of accuracy on Commonsense Reasoning benchmarks.
\end{abstract}

\section{Introduction}
\vspace{-3mm}
The increasing popularity of large generative neural networks, LLaMA (\cite{touvron2023llama}, OPT \cite{zhang2022optov}) and diffusion models (\cite{dhariwal2021diffusion,ho2020denoising}), has revolutionized the field of machine learning exhibiting exceptional capabilities in generating realistic human-like text and images. However, due to large compute and memory requirements of these models, deploying these models on resource-constrained devices is challenging. Among various compression methods available, quantization (\cite{nagel2021white,krishnamoorthi2018quantizing,jacob2018quantization}) has proven to be a promising technique to optimize various vision and language models for deployment on resource-constrained devices.

In this paper, we focus on INT4 uniform, static, channelwise weight and tensorwise activation quantization of LLMs. Most quantization literature uses dynamic or blockwise methods for 4-bit or lower quantization to recover accuracy. However, such quantization schemes may not be supported on Neural Processing Units (NPU) hardware. Channelwise weight and tensorwise activation quantization is significantly more easy to support in terms of hardware and software implementation in commercial NPUs. Keeping such strong constraints in mind, when we perform INT4 quantization of weights, quantization noise increases significantly which risks destabilizing the entire training dynamics. Therefore, we need to analyze vulnerabilities in forward and backward passes within quantized LLMs. Hence, improving accuracy for uniform, static, channelwise weight and tensorwise activation quantization is extremely important for deployment use cases of LLMs.

\begin{wrapfigure}{r}{0.5\textwidth}
    \begin{center}
               \includegraphics[width=\linewidth]{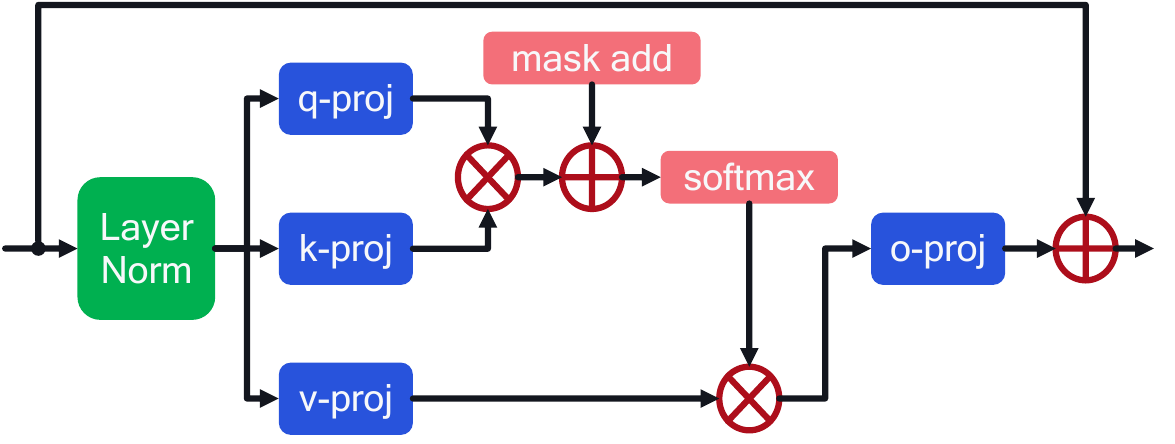}
       \caption{Multi-Head Self-Attention modules in LLaMAv2-Chat model. We focus on the forward and backward signal propagation properties after the q-, k-, v-, and o-projection layers to understand their vulnerabilities to low-bit quantization.}
       \label{fig:llama_block}
    \end{center}
\end{wrapfigure}

To this end, we propose to \textit{empirically} analyze the signal propagation through a well-known chat use case LLM, LLaMAv2-Chat, to understand its vulnerabilities to quantization errors. Specifically, we focus on multi-head self-attention modules of LLaMAv2-Chat (see Fig.~\ref{fig:llama_block}) and conduct a detailed analysis of its forward and backward pass signals to see which components of the attention modules can cause KD-QAT to destabilize. After identifying these vulnerabilities, we propose specific solutions to stabilize the KD-QAT process to achieve significant improvements in INT4 quantized LLaMAv2-Chat network accuracy enabling its deployment on commercially available hardware.

We make the following \textbf{key contributions} in this paper: 
\begin{enumerate}
    \item We create a simple pipeline for KD-QAT for a well-known chat use case LLM, LLaMAv2-Chat-7B model. It uses public datasets and requires less than a day to provide INT4 quantized networks on a single node containing 8 NVIDIA A100 GPUs. 
    \item We analyze forward and backward pass in detail for this network and find that low-bit quantization non-uniformly impacts different parts of the attention modules. Specifically, o- and v-projection layers are more susceptible to quantization errors than other weight layers in the multi-head self-attention module.
    \item Based on our analysis, we propose \textit{ov-freeze} (oh! we freeze). It shows the importance of properly analyzing layers susceptible to low-bit quantization before embarking on quantization-aware-training. We demonstrate that this improves accuracy significantly across multiple benchmarks and closes the gap between quantized and FP models.
\end{enumerate}

\label{sec:introduction}
\vspace{-0.2in}
\section{Related Work}
\vspace{-3mm}
Many methods have been proposed in the model efficiency space to solve the problem of neural network quantization. Quantization not only reduces model size but also leverages efficient fixed-point representation over floating-point representations. Most quantization techniques can be categorized either as post-training quantization (PTQ) (\cite{nagel2020up,nagel2019data,dong2019hawq}) or quantization-aware-training (QAT)(\cite{bhalgat2020lsq+,esser2019learned,nagel2022overcoming}) methods methods. Although, PTQ are go-to compression techniques (\cite{frantar2023gptq,yao2022zeroquant}) for most generative models, these methods suffer huge degradation in performance post quantization for low bitwidth quantization. QAT based approaches are difficult for these models due to lack of access to high quality data and training recipes required for finetuning these models.

In this work, we propose to use knowledge distillation to finetune weight only quantized language models to achieve near floating point performance. LLM-QAT (\cite{liu2023llm}) takes a similar approach but uses teacher generated dataset for finetuning, which is not always feasible due to sensitivity of generated dataset on sampling strategies and compute and memory intensive nature of the generation process. In this work, we use publicly available datasets for the distillation process, and perform ablations to understand vulnerabilities in the multi-head self-attention modules of the LLaMAv2-Chat model.\label{sec:Related_Work}
\vspace{-0.15in}
\section{Method}
\vspace{-3mm}
In this section, we first describe our quantization and KD-QAT setup and analyze the gradient values of the  LLaMAv2-Chat model to understand low-bit quantization vulnerabilities in the multi-head self-attention modules. Then, based on this analysis, we propose ov-freeze, a technique to stabilize our KD-QAT training pipeline. 

\subsection{Preliminaries}
While quantizing neural networks, real valued weight $W^{r}$ and activations $x^{r}$ are quantized to low precision value. Hence, for a \textit{b} bitwidth uniform quantizer with scale \textit{s} and zero-point \textit{z}, asymmetric quantization is defined as

\begin{equation}
    q(W^{r};s,z)=s \cdot \left[ clamp( \Big \lfloor\frac{W^{r}}{s}\Big \rceil + z,0,2^b-1)-z \right]  
\end{equation}

 For our experiments, we use AIMET \footnote{AIMET is a product of Qualcomm Innovation Center, Inc., available on GitHub at \url{https://github.com/quic/aimet}}  (\cite{siddegowda2022neural}) to quantize the models to desired bit-widths and use per-channel symmetric quantization for weights and asymmetric quantization for activations respectively. We set the quantization range of each weight quantizer using a mean squared error (MSE) based criteria minimizing the local MSE between FP and quantized tensors.

\subsection{KD-QAT Setup}
\vspace{-2mm}
To demonstrate the effectiveness of our KD-QAT methodology on an established chat application, we experiment with the popular LLaMAv2-Chat model (\cite{touvron2023llama}), and quantize it to W4A16\footnote{WxAy quantization indicates quantizing all the weights and output activations to x and y bits respectively.}.
For knowledge distillation based QAT, we use full-precision chat model as the teacher and quantized model as our student. Contrary to LLM-QAT (\cite{liu2023llm}), we observe that using a weighted sum of cross-entropy and KL divergence for training leads to more stable and better convergence.

\subsection{Gradient Properties of Quantized LLM}
\vspace{-2mm}
\begin{figure}[t]
  \centering
   \includegraphics[width=1\linewidth]{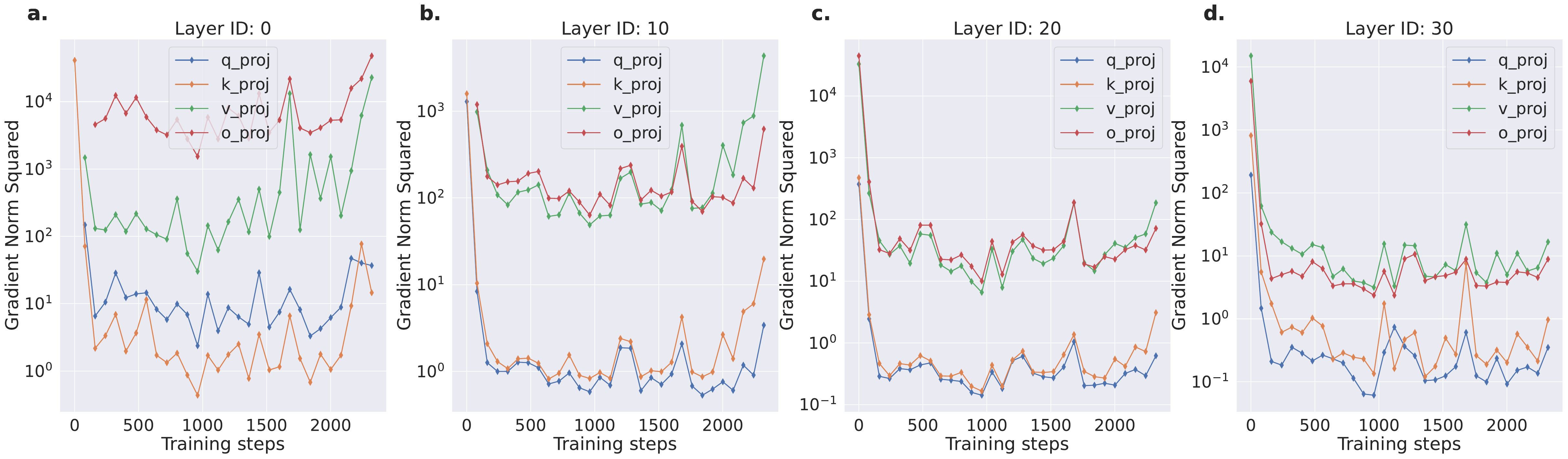}\vspace{-2.5mm}
   \caption{Relative comparison of gradient norm squared values between q-, k-, v-, o-projection outputs at various LLaMA self-attention modules during KD-QAT: (a)~Layer 0, (b)~Layer 10, (c)~Layer 20, and (d)~Layer 30. Clearly, o- and v- projection gradient norm squared values are one or two orders of magnitude higher than that of q- and k- projection layers. Such higher gradient values may destabilize o- and v-layers faster than q- and k-projections during KD-QAT, especially for low-bit quantization (e.g., INT4).}
   \label{fig:qkvograd}
\end{figure}
We investigate the gradient values of various components of the LLaMAv2-Chat multi-head self-attention modules (see Fig.~\ref{fig:llama_block}) to understand which layers may contribute to destablization when we conduct INT4 weight quantization. For layer activations with output feature map, $\Theta$, and task loss $\mathcal{L}$, we compute the frobenius norm squared of the gradients at this layer's output: 
\begin{equation}
    ||g||^{2}_2 = ||\nabla_{\Theta} \mathcal{L}(\Theta) ||^{2}_2
\end{equation}

In Fig.~\ref{fig:qkvograd}, we show the comparisons among these gradient norm squared values for q-, k-, v-, and o-projection outputs at different hidden modules\footnote{Hidden modules are labeled as Layer IDs throughout this paper. For instance, ``Layer ID: 10'' refers to the 10th hidden module out of the total 32 modules containing multi-head self-attentions in LLaMAv2 model.}. As evident, the gradient norm squared for o- and v-projections are one or two orders of magnitude higher than that for q- and k-layers. Moreover, as shown in Fig.~\ref{fig:qkvograd} (a, b), the v-layers (green lines) in early hidden modules (Layer ID: 0 and 10) seem to have largest variation in gradient norm compared to other layers. Therefore, when we perform low-bit (e.g., INT4) quantization, these observations suggest that o- and v-layers are much more vulnerable to quantization errors than q- and k- layers. Specifically, results in Fig.~\ref{fig:qkvograd} indicate that significantly higher gradient values may create bigger, more abrupt changes in o- and v- layers which would more directly impact their forward pass outputs. Therefore, forward pass of o- and v-projections may exhibit unstable behavior due to such high gradient values.

\subsection{Oh! We Freeze}
\vspace{-2.5mm}
Since o- and v-projection layers can be senstive to sudden parameter changes due to high gradients, these layers can destabilize the entire KD-QAT training trajectory. Therefore, we propose \textit{ov-freeze} to stabilize the KD-QAT training process. As the name suggests, ov-freeze fixes the o- and v-projection weights to their post-training quantization values, freezes them and trains the rest of the network. This results in model adapting to quantization error without modifying the most sensitive parts of the network. Note that, v- and o-layers constitute a much smaller part of the complete LLM since majority of the parameters are contained within the MLP layers. In the next section, we will demonstrate how the proposed method significantly stabilizes the training loop and improves the final quantization results.

\label{sec:method}
\vspace{-0.15in}
\section{Experiments}
\vspace{-3mm}
In this section, we first describe the training setup and datasets. Then, we conduct extensive experiments to show that ov-freeze stabilizes both the KD-QAT training forward and backward passes. Finally, we present results demonstrating the superior W4A16 quantized accuracy obtained using ov-freeze on several benchmark tasks.

\subsection{Training Setup and Datasets}
\vspace{-2mm}
To realize an important language use case,  on device chat applications, we perform KD-QAT experiments on the popular 7B LLaMAv2-Chat. We used a 660B-token language dataset for the finetuning process. To evaluate the performance of the quantized networks, we use Wikitext and the CommonSense benchmarks - PIQA, Arc-Easy, Arc-Challenge, WinoGrande, OpenbookQA and Hellaswag.

\begin{figure*}[t]
  \centering
   \includegraphics[width=1\linewidth]{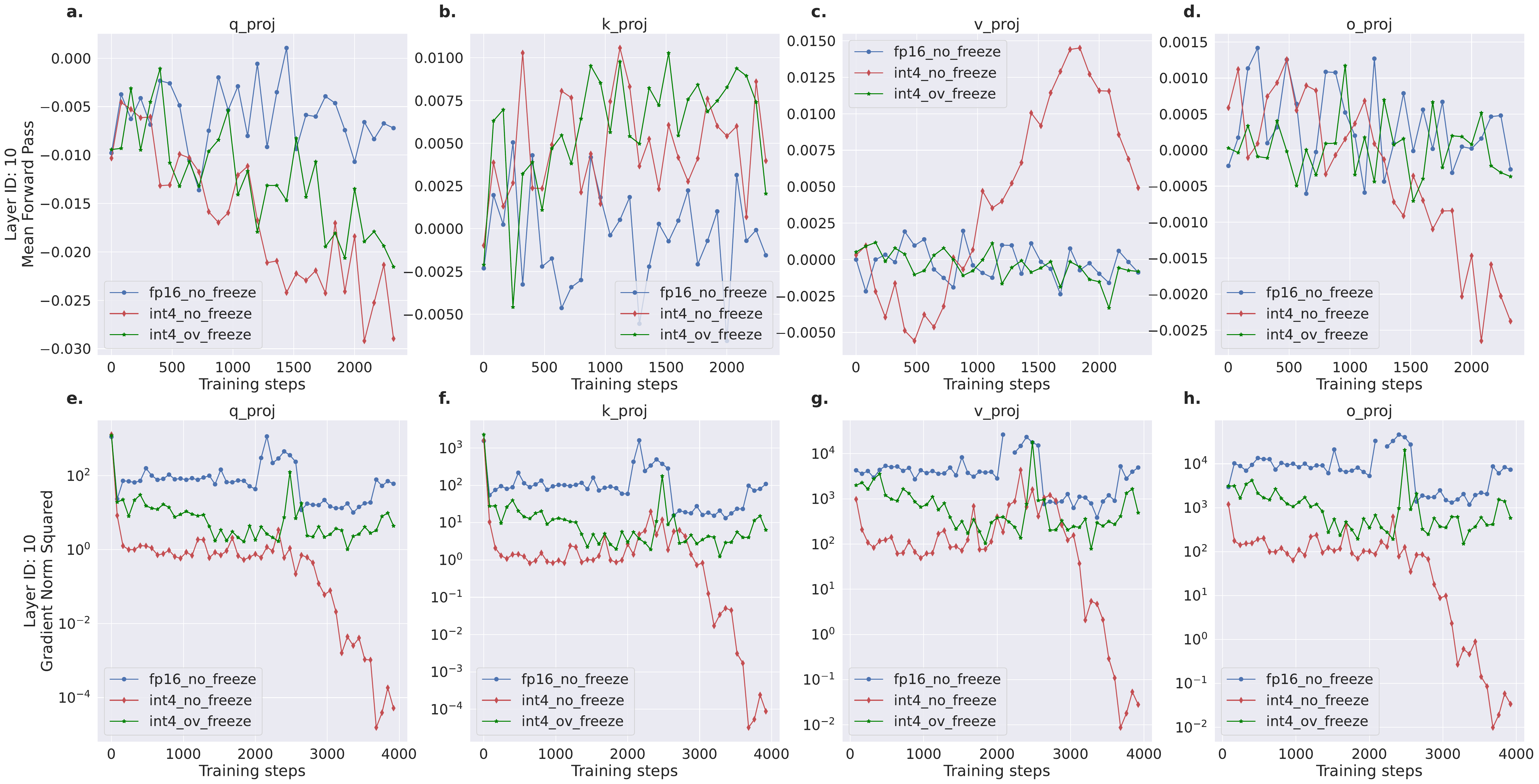}\vspace{-2mm}
   \caption{Forward and backward pass signal propagation analysis on a multi-head self-attention module at one of the hidden LLaMAV2-chat layers. We first analyze forward pass by tracking the mean value of the forward pass at the output of (a)~q-projection, (b)~k-projection, (c)~v-projection, and (d)~o-projection layers. Clearly, o and v layers are the most unstable as evident from the INT4 KD-QAT, no-freeze run. With ov-freeze, the forward pass behavior of the INT4 quantized network resembles that of the floating point model. Next, we analyze the backward pass behavior at the output of (e)~q-projection, (f)~k-projection, (g)~v-projection, and (h)~o-projection layers. Again, ov-freeze makes gradients more similar to the floating point training.}
   \label{fig:fwdbkwd}
\end{figure*}

\subsection{Signal Propagation in LLaMAv2-Chat KD-QAT}
\subsubsection{Forward Pass Analysis}
\vspace{-2mm}
As the very first metric, we analyze the mean value of output activations of certain layers in the network. Specifically, Fig.~\ref{fig:fwdbkwd} (a-d) show how the mean output activation value of q-, k-, v-, and o-projections vary across thousands of training steps for a specific LLaMAv2 hidden module. Here, we have used the FP16-no-freeze model (blue line) as a reference model. As evident from the INT4-no-freeze experiment (red line), o- and v-projection outputs deviate significantly from the FP16 reference. In contrast, when we freeze these two layers in all hidden modules (ov-freeze, green line), the mean forward pass follows the FP16 reference closely. Note that, even though we freeze only the o- and v-layers, the forward pass behavior still improves for other layers like the q-projection output shown in Fig.~\ref{fig:fwdbkwd}(a). These instabilities in forward pass can easily destabilize the training trajectory during KD-QAT. Note that, although Fig.~\ref{fig:fwdbkwd} shows results for hidden module (Layer ID) 10, we observed similar trends for other hidden modules as well (see Appendix~\ref{sec:appendix}).

Next, we empirically explore the impact of ov-freeze on backward pass.

\begin{wraptable}{r}{5.3cm}
    \addtolength{\tabcolsep}{-2pt}
    \centering
    \fontsize{6.75pt}{6.75pt}\selectfont
    
    \begin{tabular} {  lccc }
    \toprule
    \centering \textbf{Model} & \textbf{Perplexity (ppl$\downarrow$)} \\\hline
    \midrule
    FP16 & 7.08 \\
    \midrule

    PTQ: Min-Max range setting & 9.69 ({\color{red} +2.61}) \\
    PTQ: MSE-based range setting & 7.68 ({\color{red} +0.60})\\
    \midrule

    KD-QAT: Baseline (no freeze) & 7.31 ({\color{red} +0.23})\\
    KD-QAT: qkv-freeze (ours) & 7.12 ({\color{ForestGreen} +0.04})\\
    KD-QAT: v-freeze (ours) & 7.11 ({\color{ForestGreen} +0.03})\\
    KD-QAT: o-freeze (ours) & 7.09 ({\color{ForestGreen} +0.01})\\
    KD-QAT: oqkv-freeze (ours) & 6.99 ({\color{ForestGreen} --0.09})\\
    \textbf{KD-QAT: ov-freeze (ours)} & \textbf{6.98} ({\color{ForestGreen} --0.10}) \\

    \bottomrule
\end{tabular}\vspace{-2mm}

    \caption{Wikitext Perplexity for FP16 and 4-bit weight only quantized LLaMAv2-Chat Model using context length 2048 obtained through various PTQ and KD-QAT schemes. Perplexity (↓): the lower the better. {\color{red} Red} color denotes {\color{red} $>$0.2 point} and {\color{ForestGreen} Green} denotes {\color{ForestGreen} $<$0.2 point} perplexity change compared to FP16.} 
    \label{tab: wikiPPL}
\end{wraptable}
\vspace{-2mm}
\subsubsection{Backward Pass Analysis}
\vspace{-2mm}
We now analyze the gradients at the outputs of q-, k-, v-, and o-projection layers at the same hidden module from the last section. We analyze gradient norm squared values as shown in Fig.~\ref{fig:fwdbkwd} (e-h). Clearly, the INT4 ov-freeze (green line) again demonstrates similar gradient behavior as the reference FP16-no-freeze training run (blue line). In contrast, the INT4 no-freeze experiment (red line) has dramatically different backpropagation characteristics compared to the reference. All these instabilities result in lower final accuracy and an unstable KD-QAT.
\vspace{-0.1in}
\subsection{Results}
\vspace{-2.5mm}
In this section, we compare performances of FP16 and quantized models obtained by our proposed freezing scheme on multiple benchmarks. First, Table~\ref{tab: wikiPPL} reports perplexity of PTQ and KD-QAT optimized 4-bit weight only quantized LLaMAv2-Chat on the Wikitext dataset. In our ablation, we consider different freezing schemes, o-, v-, ov-, qkv- and qkvo-, and observe freezing layers with relatively higher gradient norms values, i.e. o- and v-, along with other layers help significantly in improving both the stability of the KD-QAT training and the perplexity on Wikitext, achieving at-par or better performance than the FP model. As shown in Table~\ref{tab: wikiPPL}, freezing ov- layers outperforms all other freezing schemes and PTQ baselines achieving better Wikitext perplexity than the FP16 model.

Finally, to enable on-device deployment of LLaMAv2-Chat model for practical uses, we evaluate our proposed ov-freeze W4A16 quantized model on multiple benchmarks. We use KD-QAT based weight quantized finetuned model, and quantize all activation outputs to INT16 using PTQ based min-max range setting to obtain the W4A16 model. As shown in Table~\ref{tab: CommonSense}, freezing ov-layers achieves competitive performance within 0.7\% drop of FP16 model on average on CommonSense reasoning benchmarks, thereby improving significantly over both the best MSE based RTN and AdaRound baselines, both of which lose more than 2\% accuracy compared to the FP16 network.
\begin{table*}[h]
    \addtolength{\tabcolsep}{-2pt}
    \centering
    \fontsize{7.25pt}{6.00pt}\selectfont
    \begin{tabular} {  lccccccc }
    \toprule
    \textbf{Model} & \textbf{PIQA($\uparrow$)} & \textbf{Arc-Easy($\uparrow$)} & \textbf{Arc-Challenge($\uparrow$)} & \textbf{Winogrande($\uparrow$)} & \textbf{OpenbookQA($\uparrow$)} & \textbf{Hellaswag($\uparrow$)} & \textbf{Average($\uparrow$)} \\\hline
    \midrule
    FP16 & 76.66 & 69.65 & 44.37 & 66.38 & 43.8 & 75.42 & 62.71 \\
    \midrule
    \multicolumn{8}{c}{PTQ\vspace{-0.5mm}}\\
    \midrule 
    MSE & 75.79 & 65.53 & 41.47 & 66.22 & 41.00 & 72.76 & 60.46 ({\color{red} --2.25\%}) \\
    AdaRound & 74.59 & 66.62 & 40.7 & 66.54 & 40.60 & 72.28 & 60.22 ({\color{red} --2.49\%}) \\
    \midrule
    \multicolumn{8}{c}{KD-QAT\vspace{-0.5mm}}\\
    \midrule 
    no-freeze & 76.66 & 67.38 & 42.75 & \textbf{67.17} & 41.80 & 74.14 & 61.65 ({\color{red} --1.06\%})\\
    \textbf{ov-freeze} & \textbf{77.26} & \textbf{69.11} & \textbf{42.83} & 66.54 & \textbf{42.20} & \textbf{74.15} & \textbf{62.02} ({\color{ForestGreen} --0.69\%}) \\
    \bottomrule
\end{tabular}\vspace{-3mm}
    \caption{Evaluation of FP16 and various W4A16 Quantized LLaMAv2-Chat models using 1024 context length on Commonsense Reasoning benchmarks. ($\uparrow$): the higher the better. {\color{red} Red} color denotes {\color{red} $>$1\%} and {\color{ForestGreen} Green} denotes {\color{ForestGreen} $<$1\%} loss of accuracy compared to FP16.} 
    \label{tab: CommonSense}
\end{table*}

\vspace{-2mm}

\label{sec:experiments}
\vspace{-0.2in}
\section{Conclusion}
\vspace{-2mm}
In this work, we proposed a light-weight quantization aware finetuning technique using knowledge distillation (KD-QAT) to improve the performance of popular W4A16 quantized LLM, LLaMAv2-Chat, for an important use case, on device chat applications, using commonly available datasets. We perform systematic study of the forward and backward pass of the  LLaMAv2-Chat model, and analyze the output and gradient feature maps to hint towards extreme sensitivity of o- and v-layers in the multi-head attention modules. To improve the stability of the quantized finetuning, we proposed ov-freeze and experimented with popular 7B  LLaMAv2-Chat model at 4-bit quantization level to show significant improvement in performance over vanilla QAT and achieve near float-point precision performance (within 0.7\% drop of the FP16 model) on the CommonSense reasoning benchmarks.
\label{sec:conclusion}

\bibliography{iclr2024_conference}
\bibliographystyle{iclr2024_conference}

\appendix
\section{Additional Results}\label{sec:appendix}
\begin{figure*}[h]  \centering
   \includegraphics[width=1\linewidth]{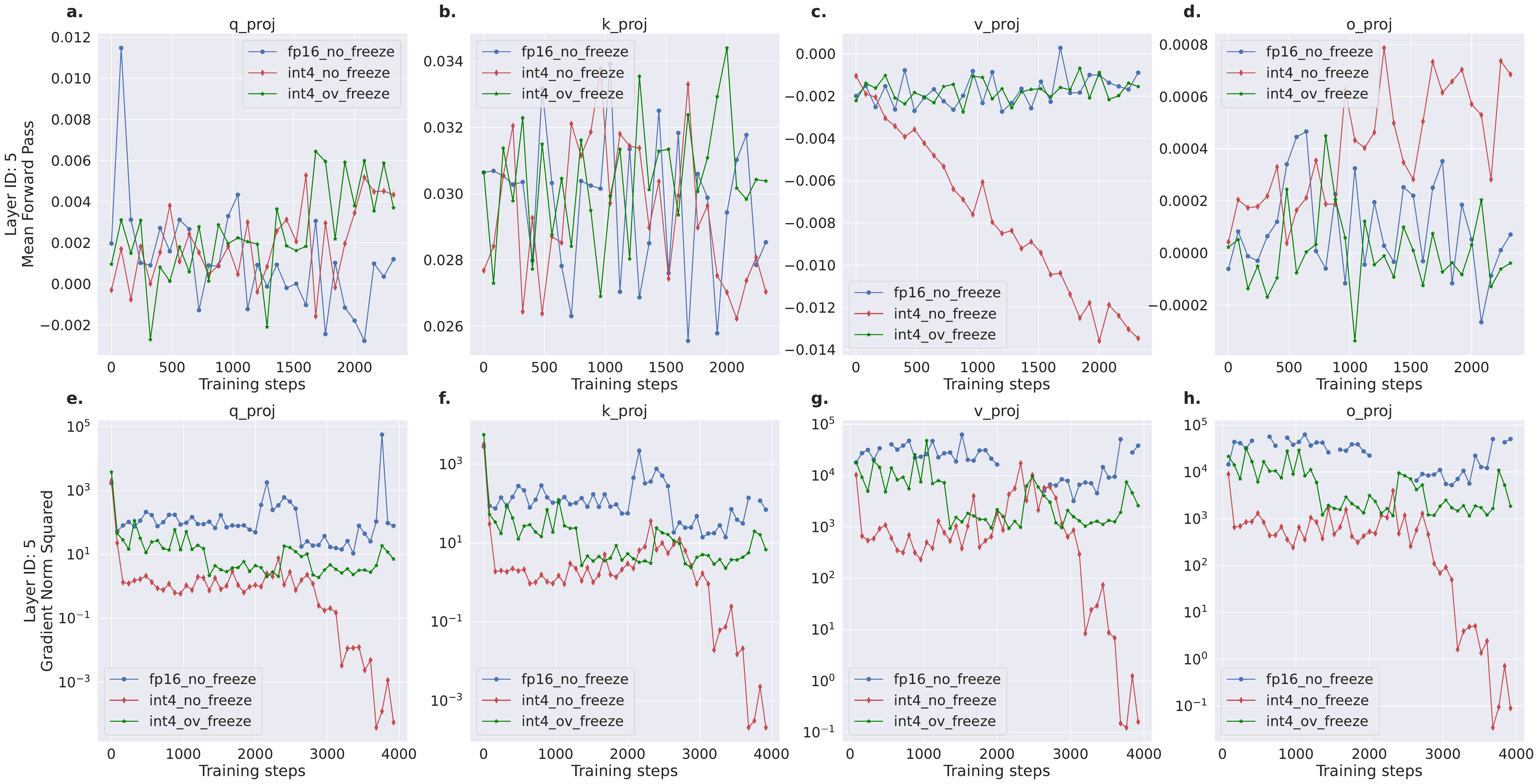}\vspace{-2mm}
   \caption{Layer ID: 5. Forward and backward pass signal propagation analysis on 5th hidden LLaMAV2-chat self-attention module. Our proposed ov-freeze makes the quantized model's forward pass and gradients more similar to those observed during FP16 training.}
   \label{fig:fwdbkwd5}
\end{figure*}

\begin{figure*}[h]  \centering
   \includegraphics[width=1\linewidth]{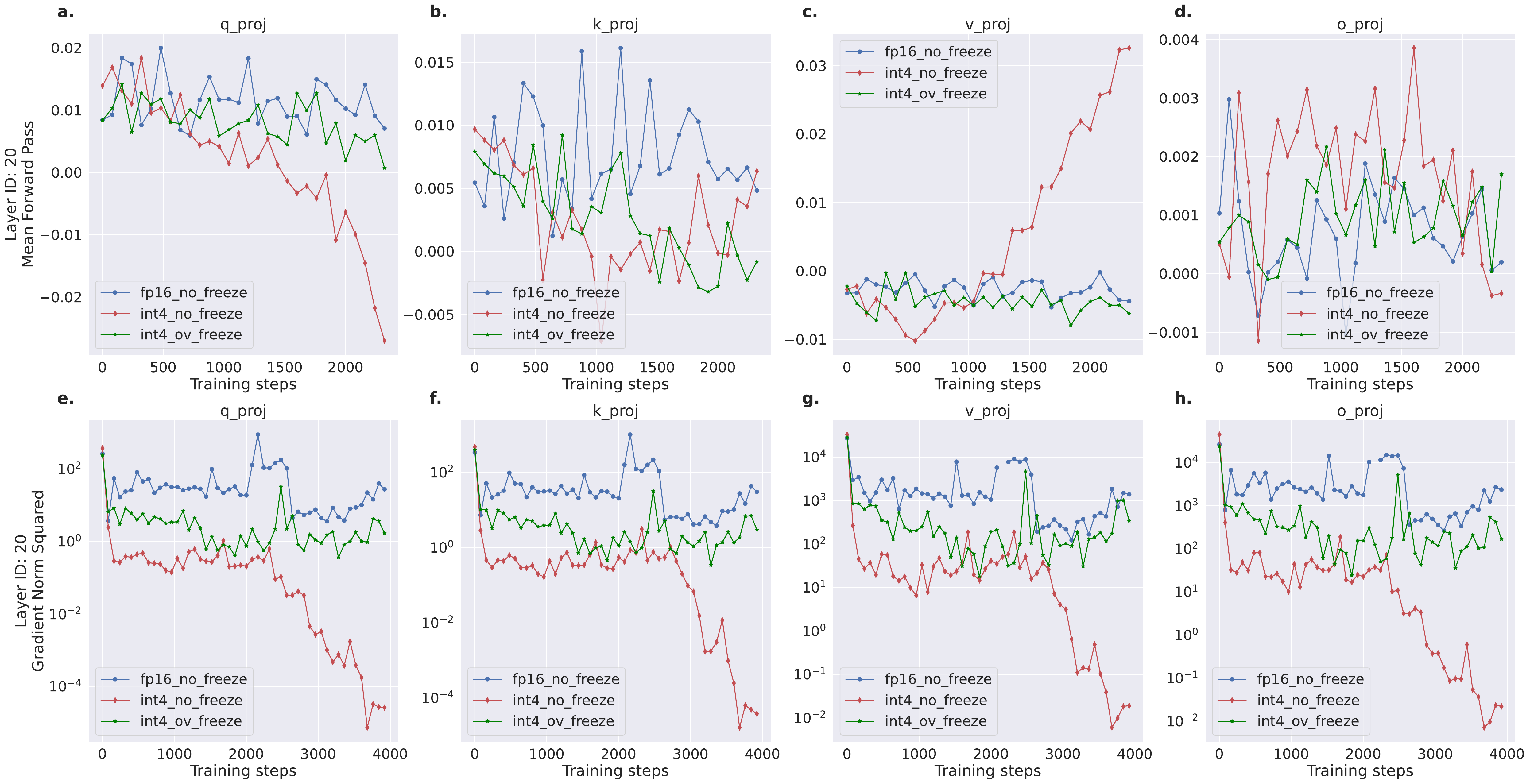}\vspace{-2mm}
   \caption{Layer ID: 20. Forward and backward pass signal propagation analysis on 20th hidden LLaMAV2-chat self-attention module. Our proposed ov-freeze makes the quantized model's forward pass and gradients more similar to those observed during FP16 training.}
   \label{fig:fwdbkwd20}
\end{figure*}

\begin{figure*}[h]  \centering
   \includegraphics[width=1\linewidth]{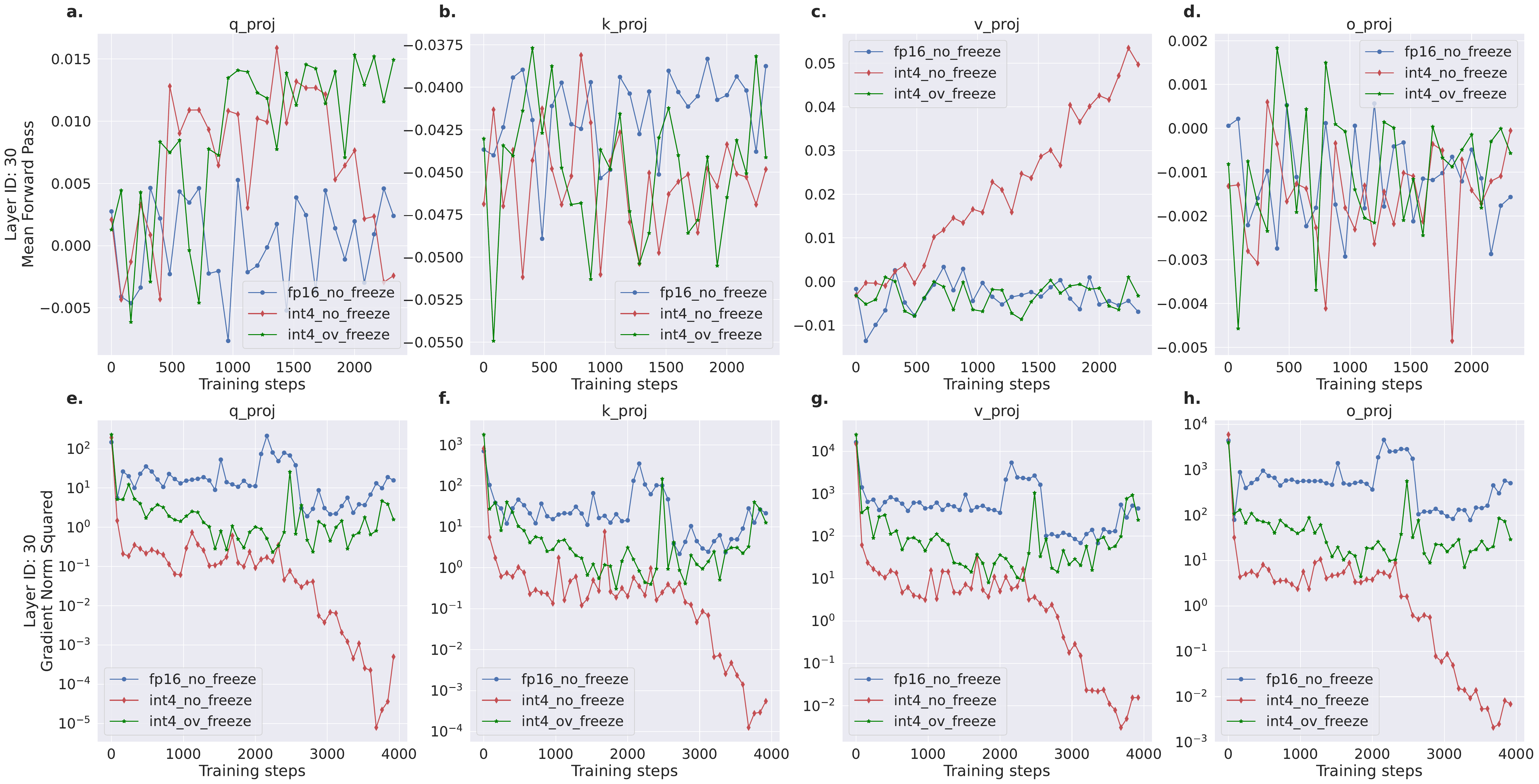}\vspace{-2mm}
   \caption{Layer ID: 30. Forward and backward pass signal propagation analysis on 30th hidden LLaMAV2-chat self-attention module. Our proposed ov-freeze makes the quantized model's forward pass and gradients more similar to those observed during FP16 training.}
   \label{fig:fwdbkwd30}
\end{figure*}

\paragraph{\textbf{Forward and Backward Signal Propagation Results on Other LLaMAv2-Chat Modules.}} Fig.~\ref{fig:fwdbkwd5}, ~\ref{fig:fwdbkwd20}, and \ref{fig:fwdbkwd30} show that our observations for hidden module 10 (i.e., Layer ID: 10 in Fig.~\ref{fig:fwdbkwd}) hold across other hidden modules. Clearly, the no-freeze mean forward pass for v-projection layer is unstable for all the hidden modules shown in Fig.~\ref{fig:fwdbkwd5}, ~\ref{fig:fwdbkwd20}, and \ref{fig:fwdbkwd30}. The proposed method ov-freeze makes the forward and backward pass characteristics similar to those seen during floating point training.

\end{document}